\pdfoutput=1

\documentclass[11pt]{article}

\usepackage[]{acl}

\usepackage{times}
\usepackage{latexsym}

\usepackage{subfig}

\usepackage[T1]{fontenc}

\usepackage[utf8]{inputenc}
\usepackage{graphicx}
\usepackage{amsmath}
\usepackage{microtype}

\title{Modeling Document-level Temporal Structures for Building Temporal Dependency Graphs}

\author{Prafulla Kumar Choubey\thanks{ ~ Work done while at Texas A\&M University} \\
  Salesforce Research \\
  \texttt{pchoubey@salesforce.com} \\\And
  Ruihong Huang \\
  Texas A\&M University \\
  \texttt{huangrh@tamu.edu} \\}

\begin{document}
\maketitle
\begin{abstract}
We propose to leverage news discourse profiling to model document-level temporal structures for building temporal dependency graphs. Our key observation is that the functional roles of sentences used for profiling news discourse signify different time frames relevant to a news story and can, therefore, help to recover the global temporal structure of a document. Our analyses and experiments with the widely used knowledge distillation technique show that discourse profiling effectively identifies distant inter-sentence event and (or) time expression pairs that are temporally related and otherwise difficult to locate\footnote{Code is available at \url{https://github.com/prafulla77/Discourse_TDG_AACL2022}}. 
\end{abstract}

\section{Introduction}
Grounding all events and time expressions to a reference timeline is fundamental to text understanding. Recently, \citet{yao-etal-2020-annotating} proposed a new task and dataset for building temporal dependency graph (TDG)\footnote{The dataset was obtained from \url{https://github.com/Jryao/temporal_dependency_graphs_crowdsourcing}}. TDG is based on the notion of narrative time and temporal anaphora, and references each timex to a timex or a meta node and each event to a timex and maybe an event. The reference timex of an event is either the smallest time (when identifiable) that encloses the event or the document creation time (DCT). Similarly, the reference event is selected such that it gives the most precise temporal interpretation for a child event. 

Because each event and timex is referenced to only one timex (or additionally an event), identified temporal relations represent the most salient relations that can potentially be used to infer additional temporal relations through transitivity or commonsense reasoning \cite{yao-etal-2020-annotating}. This makes identifying reference timex and reference event more challenging, especially when they are mentioned across sentences. Human evaluations by \cite{yao-etal-2020-annotating} also found that identifying the appropriate reference timex and reference event was the most challenging aspect of their annotation.

\begin{figure}[]
\center
{\includegraphics[width = \linewidth]{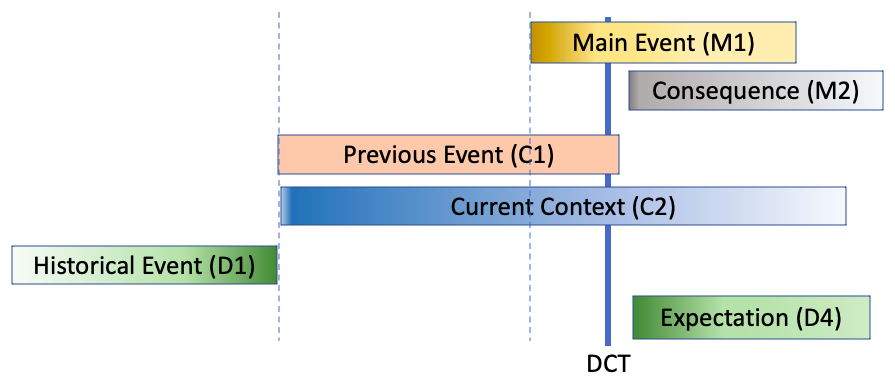}} 
  \caption{Temporal structures induced by different content types from the News Discourse Profiling.}
  \label{dp-time} 
\end{figure}

In this work, we focus on improving cross-sentence reference timex and event mentions identification by exploring discourse-level temporal cues. We choose the news discourse profiling structure (DP) \cite{choubey-etal-2020-discourse}. DP classifies sentences in a news document into one of eight content types, defined based on the functional role of a sentence in describing the main news story \cite{teun1986news,van1988news,vandijk1988news,choubey-etal-2020-discourse}, and provides an event-based functional interpretation of sentences. The eight content types include main, consequence, previous event, current context, historical, anecdotal, evaluation and expectation. 

As shown in Figure \ref{dp-time}, different content types induce different time frames relevant to a news story that can be beneficial for the global interpretation of temporal orders among event and timex mentions. For instance, mentions in historical sentences have a temporal adjacency with other mentions in historical sentences but are likely to be distant from mentions in other content types. Similarly, mentions in previous event sentences may have a temporal adjacency with mentions from one of the previous event, main event or current-context sentences but are likely to be separated from mentions in any of the historical, expectation or consequence sentences. 


We first summarize the distributional association between the position of reference mentions and discourse content types in $\S$\ref{distributional-analysis}. Then, we propose a knowledge distillation-based method to incorporate discourse knowledge into the TDG system. We experiment with the BERT \citet{devlin-etal-2019-bert} and RoBERTa \citet{liu2019roberta} pre-trained languages models and find that the proposed knowledge distillation-based TDG system is effective in using discourse-level cues and achieves improved performance on identifying cross-sentence reference mentions while retaining performance on the intra-sentence mention pairs. 

\section{Background and Analysis}
\subsection{News Discourse Profiling (DP)}
Following the news content schemata
proposed by Van Dijk \cite{teun1986news,van1988news,vandijk1988news}, DP \citep{choubey-etal-2020-discourse} defines eight content types. Each content type describes the functional role of a sentence in describing the main news event. \textit{Main event} (M1) sentence describes the major events and subjects of the news article. \textit{Consequence} (M2) describes events that are triggered by the main event. \textit{Previous Event} (C1) describes recent events that are a possible cause of the main event. \textit{Current Context} (C2) describes remaining contextual information. \textit{Historical Event} (D1) describes past events that precede the main events in months and years, \textit{Anecdotal Event} (D2)  describes unverifiable facts, \textit{Evaluation} (D3) describes opinionated contents from immediate participants, experts or journalists, and \textit{Expectation} (D4) describes speculations or possible consequences of the main or context events.

\begin{figure}[]
\center
{\includegraphics[width = \linewidth]{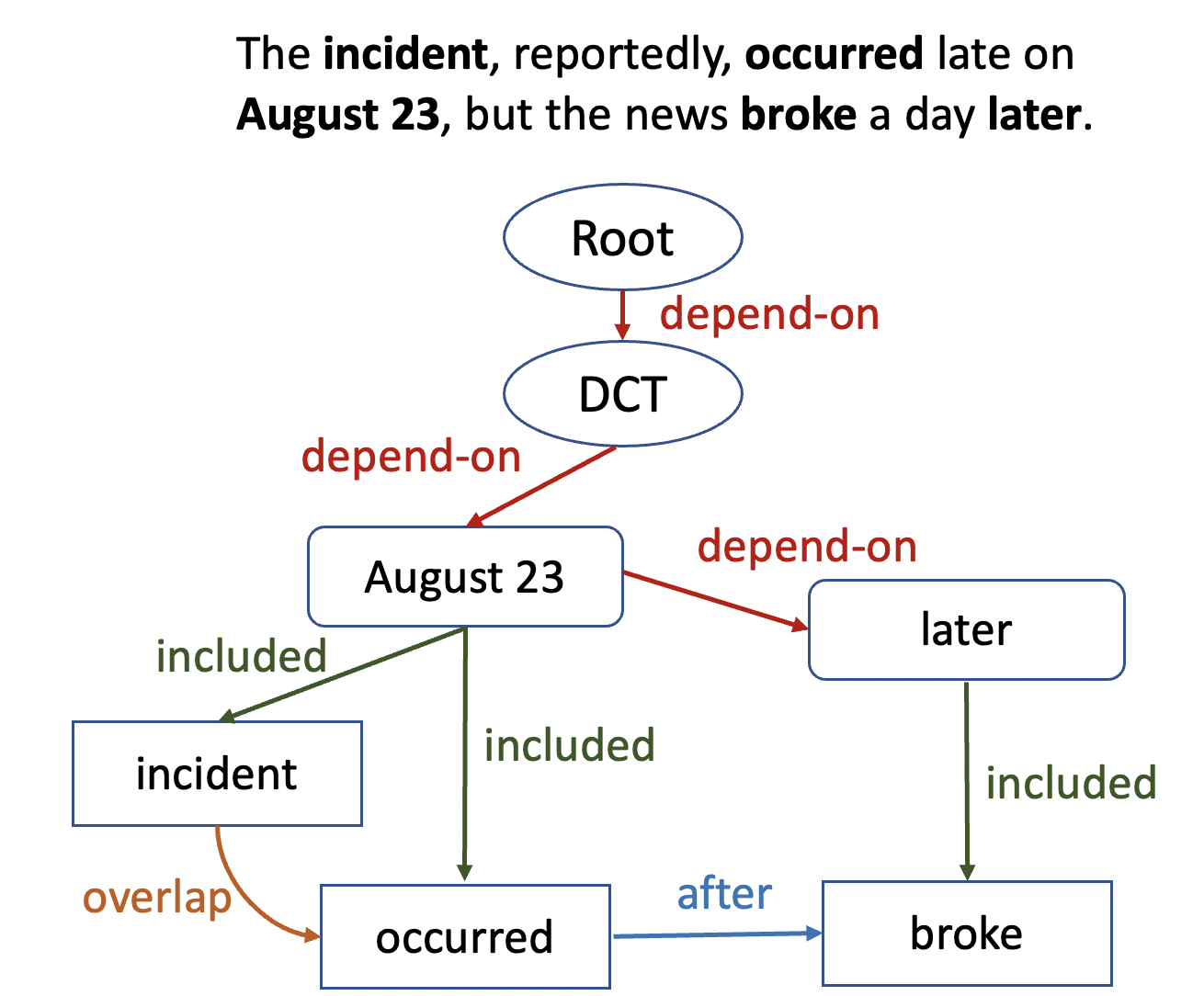}} 
  \caption{An example TDG.}
  \label{tdg-example} 
\end{figure}

\subsection{Temporal Dependency Graph (TDG)}

TDG \citep{yao-etal-2020-annotating} is a directed edge-labeled graph in which each node is either an event, a timex, or a meta node (e.g. document creation time). The reference for each timex/event node is another timex node or a meta node. Optionally, the temporal position of some events can be more precisely determined by referencing them to another event, and thus they can also have a reference event node. For instance, in Figure \ref{tdg-example}, the event \textit{incident} can only be temporally positioned with respect to the timex \textit{August 23} while the temporal order of event \textit{broke} can be determined with respect to both the timex \textit{later} and the event \textit{occurred}. The edges between event/ timex node pairs are labeled with one of the \textit{overlap}, \textit{after}, \textit{before} and \textit{included} temporal relations while the edges between a timex node and a meta node is assigned a generic \textit{depend-on} label. In this work, we focus exclusively on identifying the reference timex (and event) for each timex (event) without predicting the temporal relations between them.

\subsection{Analysis of TDG Structures w.r.t. DP Sentence Types}\label{distributional-analysis}
As illustrated in Figure \ref{dp-time}, discourse roles have temporal interpretations that are useful to locate event and timex relations in a document. Therefore, we use the recently proposed discourse profiling system by \citet{choubey-huang-2021-profiling-news}\footnote{The discourse profiling system was obtained from \url{https://github.com/prafulla77/Discoure_Profiling_RL_EMNLP21Findings}.} to assign content type labels to all sentences in the training data and analyze the distribution of reference timex and event mentions across different content types. Note that our analyses are based on a neural network model-predicted discourse content types which are noisy. Additionally, a sentence often contains more than one event and timex mentions and its content type can only provide a broad temporal ordering for constituent mentions.

First, we observe that reference timex for both timex (66\% to 100\%) and event (54\% to 80\%) mentions from all content types, except the historical, is majorly the DCT. 
Further, among the events from non-historical sentences that are not referenced to DCT, we observe that majority (71\% to 89\%) of them are referenced to a time expression from main, current-context, or previous-event sentences that overlaps with the DCT.
On the other hand, roughly 66\% of the timex mentions in historical sentences are not referenced to any timex mention but to a meta-node. Similarly, over 52\% of event mentions in historical sentences are referenced to a timex mention within the same sentence. This is expected given historical sentences describe events from the distant past that are not easily referable to current timex antecedents. 

Second, we observe that a significant proportion of cross-sentence event-event relations (45\% to 84\%) have references in either sentence of the same content type or current context sentences. This can be accounted to the anaphoric nature of TDG representation that only selects reference event which provides the most precise temporal interpretation for a given event. Since sentences with the same content types describe temporally adjacent events, they are conducive to including the most temporally salient related references for all events. 
The exact distribution of all timex and event mentions across different content types are tabulated in the appendix \ref{appendix-distribution}.

\section{Empirical Evaluations and Results}
Based on our observations in $\S$\ref{distributional-analysis}, we perform empirical evaluations to demonstrate the effectiveness of news discourse profiling for building TDG.

\subsection{Models} 
Following recent works on temporal relation identification \cite{ballesteros-etal-2020-severing} and temporal dependency parsing \cite{ross-etal-2020-exploring}, we experiment with pre-trained language models, BERT \cite{devlin-etal-2019-bert} and RoBERTa \cite{liu2019roberta}.
We model TDG as a ranking problem \cite{yao-etal-2020-annotating}, where we add a meta node each for reference timex and event. Then for each event and timex, we obtain the reference timex by selecting the one with the highest score. Similarly, we perform ranking over events to obtain the reference event for each event. To build a TDG from ranking scores, we adopt the technique used by \citet{ross-etal-2020-exploring} and iteratively select the highest-ranked reference that does not form a cycle. Within the ranking framework, we develop three models based on each of the BERT and RoBERTa to analyze the role of news discourse structure in building TDG. 

\paragraph{Baseline}: Given the sentences ($x^1_1$,..,$m_1$,..,$x^{n_1}_1$ and $x^1_2$,..,$m_2$,..,$x^{n_2}_2$) corresponding to two mentions ($m_1$  and $m_2$), we first enclose both mentions in special symbols ($\$m_1\$$ and $@m_2@$) and follow standard language model tokenization step to obtain the context representation sequence (e.g. for RoBERTa, we get $<$s$>$,$x^1_1$,..,$\$m_1\$$,..,$x^{n_1}_1$,$<$/s$>$, $<$/s$>$,$x^1_2$,..,$@m_2@$,..,$x^{n_2}_2$,$<$/s$>$). Then, we use the pre-trained model to obtain the context representation followed by a linear neural layer to obtain the final score. Note that the context sequence follows the textual order of sentences in a document.

\paragraph{DP-Feature}: In addition to the context pre-processing used for the baseline model, it appends special symbols to each sentence corresponding to its discourse content type (e.g. the context for a mention in the main sentence is represented as $x_1$,..$m_1$,..$x_n$,$\#$M1$\#$). Besides that, it mimics the baseline model.

\paragraph{DP-Distillation}: It uses the distillation technique \cite{hinton2015distilling} to introduce news discourse knowledge into our ranking system. We consider the DP model \citep{choubey-huang-2021-profiling-news} as the teacher network and the \textit{language model} component from the baseline model as the student network. The teacher model generates hard labels for sentences using the \textit{argmax} function.
Using the language model, we first obtain embeddings for all sentences in a document and then use a linear neural layer to predict their discourse content types. During training, we perform iterative gradient updates where we first update parameters based on the discourse profiling loss followed by gradient updates based on the temporal ranking loss in each batch. We observe that the order of gradient updates is important. Performing joint gradient updates or switching the order of gradient updates may significantly lower the validation performance.

\begin{table}[]
\center
\begin{tabular}{|l|c|c|} \hline
Model      & Valid & Test  \\ \hline
\citet{yao-etal-2020-annotating} & 69.0* & 79.0* \\    \hline
\multicolumn{3}{|c|}{BERT}         \\     \hline   
Baseline   &  71.90 & 76.69 \\
DP-Feature & 72.04 & 76.76 \\
DP-Distillation & \textbf{72.20} & \textbf{78.30} \\\hline
\multicolumn{3}{|c|}{RoBERTa}         \\     \hline   
Baseline   & 74.63 & 77.26 \\
DP-Feature & 74.70 & 77.30 \\
DP-Distillation & \textbf{75.03} & \textbf{78.93} \\\hline
\end{tabular}
\caption{Accuracy of different systems on the validation and test datasets. *Results for \citet{yao-etal-2020-annotating} are directly taken from the paper and correspond to the single best run. \label{result-main}}
\end{table}

\begin{table*}[]
\center
\begin{tabular}{|l|ccc|ccc|ccc|} \hline
                & \multicolumn{3}{c|}{Intra-Sentence} & \multicolumn{3}{c|}{Cross-Sentence} & \multicolumn{3}{c|}{No-Parent} \\ \hline
Model           & P          & R         & F1        & P          & R         & F1        & P        & R        & F1      \\ \hline
\multicolumn{10}{|c|}{Valid}         \\     \hline             
Baseline        & 81.03      & \textbf{84.66}     & \textbf{82.8}      & 70.60      & 65.36     & 67.86     & \textbf{70.03}    & 79.16    & 74.3    \\
DP-Distillation & \textbf{81.90}      & 83.66     & 82.76     & \textbf{72.00}      & \textbf{68.86}     & \textbf{68.76}     & 67.4     & \textbf{82.90}     & \textbf{74.33}   \\ \hline
\multicolumn{10}{|c|}{Test}  \\  \hline          
Baseline        & \textbf{80.6}       & 85.86     & 83.16     & 75.30      & 70.56     & 72.83     & \textbf{76.20}     & 80.93    & 78.43   \\
DP-Distillation & 80.53      & \textbf{86.13}     & \textbf{83.20}     & \textbf{79.90}      & \textbf{71.96}     & \textbf{75.70}     & 74.23    & \textbf{86.16}    & \textbf{79.70}  \\  \hline
\end{tabular}
\caption{Precision, recall and F1 scores for RoBERTa-based baseline and DP-distillation models on intra-sentence, cross-sentence and no-parent subsets from the validation 
datasets. \label{result-analysis}}
\end{table*}

\subsection{Experimental Settings}
We use the training, validation and test splits from \citet{yao-etal-2020-annotating} for all our experiments. Since our goal here is to evaluate the performance of a model on predicting reference timex and event mentions, we use the gold annotations for event and timex mentions. All three models are trained using AdamW optimizer \cite{loshchilov2017decoupled} for a maximum of 15 epochs and we use the epoch yielding the best validation performance. We use the batch size of 5 documents and the learning rate of 0.0001 with linear scheduling and warmup steps equivalent to 5 epochs. We search learning rate and warmup steps from [5e-4, 1e-4, 5e-6] and [3, 5, 7] respectively using the baseline model. Then, both the learning rate and warmup steps are kept constant for all models. 
Each training run takes $\sim$12 hours for the baseline and DP-feature models and $\sim$15 hours for the DP-distillation model. RoBERTa or BERT model is fine-tuned during the training. 
We run each model 3 times with random seeds and report the average performance to reduce the influence of randomness in training.

All experiments are performed on two NVIDIA-RTX-3090-24GB using PyTorch 1.7.1+cu110 \cite{NEURIPS2019_9015} and HuggingFace Transformer (v 4.0.1) libraries \cite{Wolf2019HuggingFacesTS}. We use gradient accumulation to fit a batch on 2 GPUs.

\subsection{Result and Analysis}

Table \ref{result-main} shows the results from our experiments and the previous best-performing model \cite{yao-etal-2020-annotating}. The average accuracy of the baseline model, which relies on the pre-trained RoBERTa (BERT), is 5.63\% (2.90\%) higher than the best performing neural model from \citet{yao-etal-2020-annotating} on the validation dataset. Surprisingly, on the test dataset, our RoBERTa (BERT)-based baseline model achieves 1.74\% (2.31\%) lower average accuracy. 

Next, using discourse content types as a feature in the input sequence brings negligible improvement over the baseline for both RoBERTa and BERT-based models. We suspect that special symbols used to represent each content type are unaware of the temporal associations between different content types. Thus, the DP-feature model is only capable of modeling co-occurrences of different content types with reference event and timex mentions. Additionally, the pre-training of the BERT/ RoBERTa model did not consider special content types symbols which leads to inconsistent interpretation of their corresponding tokens during the pre-training and the fine-tuning steps.

DP-distillation method using the RoBERTa (BERT) model, on the other hand, improves the average accuracy of Baseline by 0.4\% (0.3\%) and 1.66\% (1.61\%) on validation and test datasets respectively. Training with the distillation technique enables the transfer of DP knowledge directly from the teacher DP model into the student RoBERTa/ BERT model, unlike the DP-feature model which is unaware of DP knowledge unless specified through features. Further, the DP-distillation model learns to predict content type labels, while being validated over performance on ranking true reference mentions, which provides it with higher flexibility to distill and retain directly relevant knowledge.

\paragraph{Why Discourse Profiling helps?} 
Since DP provides temporal cues at the sentence level, we mainly expect the performance improvement to come from cross-sentence event/ timex pairs. To verify that, we partition our validation and test datasets into three subsets: 1) \textit{intra-sentence} that includes pairs with both given mention and reference mention from the same sentence, 2) \textit{cross-sentence} that includes pairs with given mention and reference mention from different sentences, and 3) \textit{no-parent} that includes mentions which are referenced to a meta node. 
We compare the RoBERTa based-baseline and DP-distillation models, which perform better than the corresponding BERT-based models, on three data partitions in Table \ref{result-analysis}.
As expected, we found that both the baseline and DP-distillation models achieve comparable performance on the same-sentence subset. For the no-parent subset, we observe higher recall and lower precision for the DP-distillation model. Intuitively, the model learns to link more event and timex mentions to a meta node. Note that timex mentions from historical sentences are majorly linked to a meta node ($\S$\ref{distributional-analysis}), which may be partly responsible for this behavior.

On the cross-sentence subset, we observe consistent improvement on all precision, recall and F1 scores for the DP-distillation model. This is consistent with our hypothesis that discourse profiling can be used to induce document-level temporal structures and help in identifying references for event/timex mentions that require cross-sentence temporal cues.

\section{Related Work}
Most previous works \cite{mani-etal-2006-machine,bethard-martin-2007-cu,kolomiyets-etal-2012-extracting,dsouza-ng-2013-classifying,bethard-2013-cleartk,ng-etal-2013-exploiting,laokulrat-etal-2013-uttime,mirza-tonelli-2014-classifying,choubey-huang-2017-sequential,yao-etal-2017-weakly,dai-etal-2017-using,yao-huang-2018-temporal,ballesteros-etal-2020-severing} treat temporal relation extraction as a pair-wise classification problem and most widely used datasets follow the same pair-wise schema for annotating temporal relations between event/ timex pairs \cite{article_acquaint,timebank2003,pustejovsky2003timeml,cassidy-etal-2014-annotation,uzzaman-etal-2013-semeval,ning-etal-2018-multi}. However, as discussed by \citet{zhang-xue-2018-structured,zhang-xue-2018-neural,ross-etal-2020-exploring,yao-etal-2020-annotating}, pairwise annotations as well as classification models suffer from quadratic complexity, partial annotations and inconsistent predictions.
Recently, \citet{zhang-xue-2018-structured} proposed to build a dependency tree (TDT) structure to address the above three problems with pair-wise annotations and modeling and later extended that to temporal dependency graph \cite{yao-etal-2020-annotating}. We use the most recent temporal dependency graph dataset that improves the expressiveness of previous TDT datasets \cite{zhang-xue-2018-structured,zhang-xue-2019-acquiring} and follow their neural ranking modeling approach. However, different from the previous work, we explore news discourse profiling to explicitly focus on improving the performance of a neural ranking model on cross-sentence event/ timex pairs.

\citet{ng-etal-2013-exploiting} were the first to show the effectiveness of several discourse analysis frameworks, including rhetorical structure theory (RST) \cite{mann1988rhetorical}, PDTB-style discourse relations \cite{prasad2008penn} and topical text segmentation \cite{hearst-1994-multi} for temporal relation extraction. Different from the above three discourse structures, discourse profiling is a functional \cite{webber-joshi-2012-discourse} structure and has global event-centric interpretations. Secondly, \citet{ng-etal-2013-exploiting} focused on classifying temporal relations between a given pair of temporally related events. In contrast, our goal is to identify the most salient reference for every event/ timex mention that determines its most precise location on the timeline. 

\section{Conclusion}
We have shown that news discourse profiling can be used to incorporate document-level temporal structures when building temporal dependency graphs. Through analyses, we have shown the distributional association between discourse content types and positions of reference and child mentions. Further, empirical evaluation using the knowledge distillation technique shows that discourse profiling is effective in identifying cross-sentence reference-child mention pairs. In the future, we will explore new linguistics structures and modeling techniques to incorporate document-level temporal structures for building TDG. 

\section{Acknowledgements}
We gratefully acknowledge support from National
Science Foundation via the awards IIS-1942918. We would also like to thank the anonymous reviewers for their feedback.

\bibliography{anthology,custom}
\bibliographystyle{acl_natbib}

\appendix

\section{Responsible NLP Research Checklist}
\subsection{Limitations and Risks}
Our proposed method relies on news articles' specific functional discourse structure, called news discourse profiling. This limits the applicability of the method to the news domain only. We run all experiments on the dataset in the English language. While we expect the method to work well for other languages, provided we have a dataset/ model for constructing the news discourse profiling structure in the target language, we have not verified this experimentally. Our results are based on the average of 3 runs with random seeds. We do not expect any potential risk from the proposed method.

\subsection{Artifacts}
We use two publicly available datasets, TDG corpus \cite{yao-etal-2020-annotating} and NewsDiscourse corpus \cite{choubey-etal-2020-discourse}, for our experiments and analyses. Our implementations are based on the HuggingFace transformers \cite{Wolf2019HuggingFacesTS} (Apache license 2.0) and we will release our code under the BSD 3 license.

\section{Distributional Analysis} \label{appendix-distribution}

\begin{table}[!h]
\center
\begin{tabular}{|l|cc|} \hline
   & DCT   & Meta-node \\ \hline
M1 & \textbf{86.5}  & 8.5       \\
M2 & \textbf{88.9}  & 4.4       \\
C1 & \textbf{81.9}  & 9.0       \\
C2 & \textbf{79.8}  & 14.6      \\
D1 & 30.9  & \textbf{66.1 }     \\
D2 & \textbf{100.0} &  -         \\
D3 & \textbf{88.8}  & 8.8       \\
D4 & \textbf{88.4}  & 10.9      \\
NA & \textbf{66.7}  & 25.0\\ \hline
\end{tabular}
\caption{Distribution of timex and their reference timex mentions, for each content type.}
\end{table}

\begin{table}[!h]
\center
\begin{tabular}{|l|cc|} \hline
   & DCT   & Intra-sentence \\ \hline
M1 & \textbf{58.4} & 30.4          \\
M2 & \textbf{60.1} & 17.5          \\
C1 & \textbf{54.3} & 28.7          \\
C2 & \textbf{63.6} & 17.1          \\
D1 & 34.4 & \textbf{52.3}          \\
D2 & \textbf{73.5} & 6.0           \\
D3 & \textbf{80.8} & 7.1           \\
D4 & \textbf{75.6} & 15.8          \\
NA & \textbf{69.0} & 20.0         \\ \hline
\end{tabular}
\caption{Distribution of event and their reference timex mentions, for each content type.}
\end{table}

\begin{table*}[!h]
\center
\begin{tabular}{|l|ccccccccc|} \hline
   & M1   & M2  & C1   & C2   & D1   & D2  & D3  & D4   & NA  \\ \hline
M1 & 39.8 & 1.1 & 17.2 & 29.0 & 6.5  &  -   & 1.1 & 2.2  & 3.2 \\
M2 & 69.7 & 4.5 & 7.6  & 13.6 & 4.5  &  -   &   -  &   -   &   -  \\
C1 & 31.7 &  -   & 43.9 & 10.1 & 9.4  & 1.0 & 3.6 & 1.0  &  -   \\
C2 & 36.8 & 1.1 & 17.1 & 29.2 & 9.3  &  -   & 3.5 & 1.8  & 1.3 \\
D1 & 7.2  &  -   & 39.6 & 10.8 & 36.0 &  -   & 4.5 &  -    & 1.8 \\
D2 & 61.3 &  -   & 9.7  & 19.4 & 6.5  &  -   & 3.2 &   -   &  -   \\
D3 & 27.0 & 1.9 & 17.1 & 31.4 & 8.8  & 1.0 & 5.0 & 3.6  & 4.7 \\
D4 & 39.7 & 2.9 & 10.3 & 22.1 & 5.9  &  -   & 4.4 & 11.8 & 2.9 \\
NA &  -    &  -   &   -   & 81.8 & 18.2 &  -   &  -   &   -   &   - \\ \hline
\end{tabular}
\caption{Distribution of event and their reference timex mentions over different content types, when the reference timex is not the DCT. We can see that majority (71\% to 89\%) of
the events from non-historical sentences are referenced to a time expression from main,
current-context, or previous-event sentences that overlaps with the DCT.}
\end{table*}

\begin{table*}[!h]
\center
\begin{tabular}{|l|ccccccccc|} \hline
   & M1   & M2   & C1   & C2   & D1   & D2  & D3   & D4   & NA  \\  \hline
M1 & 43.9 & 1.8  & 10.5 & 21.1 & 5.3  &   -  & 12.3 & 1.8  & 3.5 \\
M2 & 19.6 & 17.4 & 6.5  & 52.2 & 2.2  &   -  &  -    & 2.2  &   -  \\
C1 & 15.8 & 1.7  & 35.0 & 30.8 & 10.0 & 0.8 & 3.3  & 1.7  & 0.8 \\
C2 & 8.7  & 2.8  & 7.9  & 62.1 & 5.6  & 0.2 & 9.2  & 3.3  & 0.1 \\
D1 & 3.4  & 0.7  & 4.7  & 31.8 & 46.6 &  -   & 10.8 & 2.0  &   -  \\
D2 & 57.6 & 3.0  &   -   & 15.2 & 3.0  &   -  & 18.2 &   -   & 3.0 \\
D3 & 1.7  & 0.6  & 3.5  & 30.3 & 3.5  & 0.6 & 54.1 & 4.5  & 1.1 \\
D4 & 5.9  & 1.7  & 5.9  & 33.1 & 6.8  &  -   & 33.1 & 12.7 & 0.8 \\
NA &   -   & 0.1  &   -   & 0.2  & 0.1  & 0.1 & 0.3  &   -   & 0.3 \\  \hline
\end{tabular}
\caption{Distribution of cross-sentence event and their reference event mentions over different content types, where the event and its reference event are from different sentences. We can see that significant proportion of cross-sentence event-event temporal links (45\% to 84\%) have references in either sentences of the same content type or current context sentences.}
\end{table*} 

\end{document}